\documentclass{article}

\usepackage{arxiv}

\usepackage[utf8]{inputenc} 
\usepackage[T1]{fontenc}    
\usepackage{hyperref}       
\usepackage{url}            
\usepackage{booktabs}       
\usepackage{amsfonts}       
\usepackage{nicefrac}       
\usepackage{microtype}      
\usepackage{lipsum}
\usepackage{subfig}
\usepackage{graphicx}

\title{Virtual to Real adaptation of Pedestrian Detectors}

\author{
  Luca Ciampi\thanks{Institute of Information Science and Technologies - National Research Council - Pisa, Italy. \newline Corresponding authors: luca.ciampi@isti.cnr.it, nicola.messina@isti.cnr.it} 
   \And
 Nicola Messina$^{*}$
   \And
 Fabrizio Falchi$^{*}$
     \And
 Claudio Gennaro$^{*}$
     \And
 Giuseppe Amato$^{*}$
}

\begin{document}
\maketitle

\newcommand{\ourdataset}{\textit{ViPeD }}
\newcommand{\ourdatasetwithoutspace}{\textit{ViPeD}}

\begin{abstract}
Pedestrian detection through Computer Vision is a building block for a multitude of applications. Recently, there has been an increasing interest in convolutional neural network-based architectures to execute such a task. One of these supervised networks' critical goals is to generalize the knowledge learned during the training phase to new scenarios with different characteristics. A suitably labeled dataset is essential to achieve this purpose. The main problem is that manually annotating a dataset usually requires a lot of human effort, and it is costly. To this end, we introduce {\ourdataset} ({Vi}rtual {Pe}destrian {D}ataset), a new synthetically generated set of images collected with the highly photo-realistic graphical engine of the video game {GTA V (Grand Theft Auto V}), where annotations are automatically acquired. However, when training solely on the synthetic dataset, the model experiences a {Synthetic2Real} domain shift leading to a performance drop when applied to real-world images. To mitigate this gap, we propose two different domain adaptation techniques suitable for the pedestrian detection task, but possibly applicable to general object detection. Experiments show that the network trained with \ourdataset can generalize over unseen real-world scenarios better than the detector trained over real-world data, exploiting the variety of our synthetic dataset. Furthermore, we demonstrate that with our domain adaptation techniques, we can reduce the Synthetic2Real domain shift, making the two domains closer and obtaining a performance improvement when testing the network over the real-world images.
\end{abstract}


\section{Introduction}
\label{sec:introduction}
A key task in many intelligent video surveillance systems is pedestrian detection, as it provides essential information for semantic understanding of video. Accurate detection of individual instances of pedestrians in images plays a vital role in a myriad of applications that can positively impact the quality of human life. They range from video surveillance \cite{bilal2016low, varga2017robust}, robotics, automotive \cite{gavrila2007multi, shashua2004pedestrian} and assistive technologies to people with visual disabilities \cite{tian2014rgb}, just to name a few. 

Convolutional neural networks-based methods \cite{lecun1998gradient} have recently demonstrated their superiority compared to the approaches relying on hand-crafted features. However, despite the recent advances, the pedestrian detection task remains a challenging active research area in Computer Vision. While there exist some large annotated generic datasets suitable for training these supervised learning networks, such as {ImageNet} \cite{Deng2009} and {MS COCO} \cite{coco}, in many real-world situations they are not enough. Hence, as a consequence, a model trained using these data usually experiences a drastic drop in performance when applied to another scenario at inference time.

The crux of Convolutional Neural Networks (CNNs) 
 is that, to generalize well at inference time, they require a huge amount of diverse labeled data during the training phase, covering the widest number of different scenarios. Since manually annotating new collections of images is expensive and requires a great human effort, a recently promising approach is to gather data from virtual world environments that mimics as much as possible all the characteristics of the real-world scenarios, and where the annotations can be acquired with a partially automated process. To this end, in this work, we provide {\ourdataset} ({Vi}rtual {Pe}destrian {D}ataset), a new {synthetic} dataset generated with the highly photo-realistic graphical engine of the video game {GTA V (Grand Theft Auto V)} by {Rockstar  North}, that extends the {JTA (Joint Track Auto)} dataset presented in \cite{Fabbri2018}.

The use of synthetic datasets based on 3D rendering to tackle the annotation problem is not new. Some notable examples are GTA5 \cite{richter2016playing} and SYNTHIA \cite{ros2016synthia} for semantic segmentation. However, to the best of our knowledge, {\ourdataset} is the first synthetic dataset suitable for the pedestrian detection task, which is annotated with bounding boxes locating the people's instances present in the scenes.

While synthetic data collections are very appealing, usually, when training solely on a synthetic dataset, the model does not generalize well to real-world data. This performance gap is due to the fact that the network learns from one domain (named {training} or {source} domain) and is then applied on another different domain ({test} or {target} domain), and is commonly referred as Domain Shift \cite{torralba2011unbiased}. In this particular case, the source and the target domains are the synthetic and the real-world ones, respectively. Hence, we call this Domain Shift as {Synthetic2Real}.


In this paper, we propose two different {Domain Adaptation} (DA) methods to mitigate this {Synthetic2Real} Domain Shift, suitable for the pedestrian detection task, but possibly applicable to general object detection. The first one consists of training the model exploiting the synthetic data, and then, in a second step, fine-tuning it using the real-world images. Instead, the second one consists of an end-to-end training procedure in which we employ mixed batches containing both synthetic and real data.

First, we test the generalization capabilities of the detector over unseen scenarios. We show that we can obtain better or comparable results when training exploiting the synthetic data than when using the same model trained using only real-world images, just taking advantage of the variety of {\ourdatasetwithoutspace}. Secondly, we experiment with the two proposed domain adaptation techniques to boost the performance over specific real-world scenarios. We demonstrate that we can reduce the {Synthetic2Real} Domain Shift by bringing the two domains closer together, thus achieving better results. 

Summarizing, the main contributions of this work are the followings:
\begin{itemize}
    \item We introduce and make publicly available {\ourdatasetwithoutspace}, a new vast synthetic dataset suitable for the pedestrian detection task, generating the images using photo-realistic video game {GTA V (Grand Theft Auto V)}, that extends the {JTA (Joint Track Auto)} dataset presented in \cite{Fabbri2018}. 
    \item We present two supervised Domain Adaptation techniques to mitigate the {Synthetic2Real} Domain Shift existing between the synthetic and the real images. 
    \item We conduct extensive experimentation on various real-world pedestrian detection datasets present in the literature. First, we test the detector's generalization capabilities, demonstrating that we achieve comparable or better results using synthetic data during the training phase rather than relying solely on the real-world images. Second, we experiment with the two proposed DA solutions to boost the performance over specific real-world scenarios, bringing the synthetic and the real domains closer, achieving better results.
\end{itemize}

Specifically, in this work, we extend our previous paper \cite{amato2019viped}. Compared to \cite{amato2019viped}, we obtain better results, employing a new state-of-the-art detector that exhibits higher performance and introducing a new domain adaptation strategy. Furthermore, we carry out extensive experimentation over additional publicly available datasets, demonstrating the robustness of our approach over different real-world scenarios. The code, the models, and the dataset are made freely available at \href{https://ciampluca.github.io/viped/}{https://ciampluca.github.io/viped/}.

\section{Related Work}
In this section, we review some relevant works about the object and pedestrian detection task. We also analyze some previous studies on DA, focusing on the {Synthetic2Real} domain shift.

\subsection{Pedestrian Detection}
Pedestrian detection is highly related to object detection. It deals with locating and recognizing instances of pedestrians' specific class, usually in images of urban environments, without taking into account group dynamics. We can subdivide approaches for the pedestrian detection task into two main research areas. The first class of detectors is based on handcrafted features, such as ICF (Integral Channel Features) \cite{10yearspedestrian, Zhang2014, Zhang2015, Zhang2016, Nam2014}. These methods can usually rely on higher computational efficiency, at the cost of lower accuracy. On the other hand, more recently, Deep Neural Network approaches have been explored. For example, \cite{Tian2015, Yang2016, Cai2016, Sermanet2013} proposed some solutions based on the CNN networks \cite{lecun1998gradient} to detect pedestrians, even accounting for different scales.

Recent advances using CNNs were also possible thanks to the availability of many new datasets. Some of the most used in literature are {Caltech} \cite{Dollar2012}, {INRIA} \cite{Dalal2005}, {MOT17Det} \cite{MOT17}, {MOT19Det} \cite{MOT19}, and {CityPersons} \cite{citypersons}. 
In this work, we considered the latter three ones since they describe very heterogeneous video-surveillance scenarios, and they have proved to be enough challenging due to their high variability outlining most of the real-world problematic situations. The {Caltech} and the {INRIA} datasets are instead specifically collected for detecting pedestrians in self-driving contexts, a different scenario not considered in this paper.

\subsection{Synthetic2Real Domain Adaptation}
With the need for huge amounts of labeled data, synthetically-generated datasets have recently gained considerable interest. Some notable examples are GTA5 \cite{richter2016playing} and SYNTHIA \cite{ros2016synthia} for semantic segmentation.  

However, as already mentioned in the introduction, there is a non-negligible domain gap between the synthetic and the real worlds. Many techniques try to fill this gap, using both supervised and unsupervised approaches. An exhaustive survey about deep learning DA techniques is provided in \cite{wang2018deep}. For example, authors in \cite{ciampi2020unsupervised} and in \cite{tsai2018learning} proposed two unsupervised domain adaptation solutions for the counting task and the segmentation task, respectively, taking advantage of the output space. Authors in \cite{Fabbri2018} created JTA (Joint Track Auto), a synthetic dataset taken from the highly photo-realistic video game GTA V. They demonstrated that it is possible to reach excellent results on tasks such as people tracking and pose estimation when validating on real data. In this work, we extend this dataset, making it suitable for the pedestrian detection task.

Authors in \cite{Kaneva2011, Marin2010} have also focused on learning features from synthetic data for the pedestrian detection task. Still, they did not take into account deep learning approaches, exploring only traditional detection techniques. In \cite{bochinski2016training}, instead, the authors employed a synthetic dataset to train a CNN able to detect objects belonging to different classes in a video. This CNN is responsible only for the classification of the objects, while the detection of them relied on a background subtraction algorithm based on Gaussian Mixture Models (GMMs). This approach's performance over real-world scenarios was evaluated employing two pedestrian detection datasets, one of which, the 2D MOT 2015 \cite{mot2015}, is an older version of the dataset we used to carry out our experiments. To the best of our knowledge, \cite{Roberson2016} and \cite{bochinski2016training} are the closest works to our. In particular, they also used {GTA V} as the source for the acquisition of the synthetic data, but they focus their efforts on the vehicle detection task.

\section{\ourdataset ({Vi}rtual {Pe}destrian {D}ataset)}
In this section, we illustrate the motivations for using synthetic data, pointing out the main benefits and drawbacks. Then, we introduce and describe the construction of {\ourdataset}, our synthetic collection of images exploited for training the pedestrian detector.

\subsection{Training with Synthetic Datasets}
As already pointed out in Section \ref{sec:introduction}, the main drawback of CNN-based methods is that they hinge on large quantities of annotated data. Since they require ground truth labels for supervised learning, they may not generalize well to unseen images, especially when there is a large domain gap between the training (source) and the test (target) sets, such as different perspectives, illuminations, and object scales. This gap often severely hampers the application of CNN-based solutions to very large scale scenarios since annotating images for all the possible cases is an expensive operation, implying a considerable human-effort.

A possible solution to this problem is to create a vast and suitable dataset by collecting images from virtual world environments that resemble, as closely as possible, all the characteristics of the target real-world scenarios. Here, the main advantage is that the labels of the images can be acquired with a partially automated process, and so the data collection is significantly less costly. Consequently, it is possible to record a considerable amount of images covering a large number of different scenarios.

However, besides these positive aspects, there are some drawbacks to be considered. In particular, synthetic images' appearance is still significantly different from that observed in real-world images, even using current rendering techniques. Thus, the model trained solely on the synthetic dataset does not generalize to real-world data as one might expect due to the {Synthetic2Real} Domain Shift. 

With the purpose of reducing the described domain shift, domain adaptation techniques can be exploited during the CNN-based networks' training phase. These methods try to make more similar the two data distributions, i.e., the distribution of the features belonging to the synthetic world and the one belonging to the real-world environment. Thus, it is possible to take advantage of the synthetic dataset's diversity, mitigating the underlying differences between the two domains.

\subsection{ViPeD}
\label{subsec:viped}
{\ourdataset} ({Vi}rtual {Pe}destrian {D}ataset) is a new {synthetic} dataset generated with the highly photo-realistic graphical engine of the video game {GTA V (Grand Theft Auto V)} by {Rockstar North}. It extends JTA (Joint Track Auto) dataset, presented in \cite{Fabbri2018}. The dataset includes a total of about 500K images, extracted from 512 full-HD videos of different urban scenarios. These videos are organized into a training set (256 videos) and a test set (the remaining 256 videos).


While we can reuse the already existing JTA images, we need to generate suitable annotations for the pedestrian detection task. Indeed, the JTA dataset provides only skeletal information useful in the pose estimation and tracking tasks. In our scenario, instead, we are required to annotate each pedestrian with the four coordinates (x, y, w, h) delimiting its minimum enclosing bounding box. Hence, we employed the already available JTA images producing a new set of labels suitable for our task.


Estimating the precise bounding box surrounding each pedestrian instance can be tricky, as we do not have access to the underlying GTA game engine. Other works tried to overcome this problem by using some interesting work-around. For example, \cite{Roberson2016} extracted the semantic masks around each object in the scene and separated the instances by exploiting the depth information available through the depth buffer.  

Our solution relies instead on the skeletal information already provided by the JTA annotations. Indeed, differently from \cite{Roberson2016}, we deal with multiple instances of pedestrians in possibly highly crowded scenarios. In these cases, the depth information may be insufficient for distinguishing two different pedestrians, leading to possible severe bounding box estimation errors.

\begin{figure}[htbp]
    \centering
  \subfloat[\label{dataset_skeleton}]{%
       \includegraphics[width=0.46\linewidth]{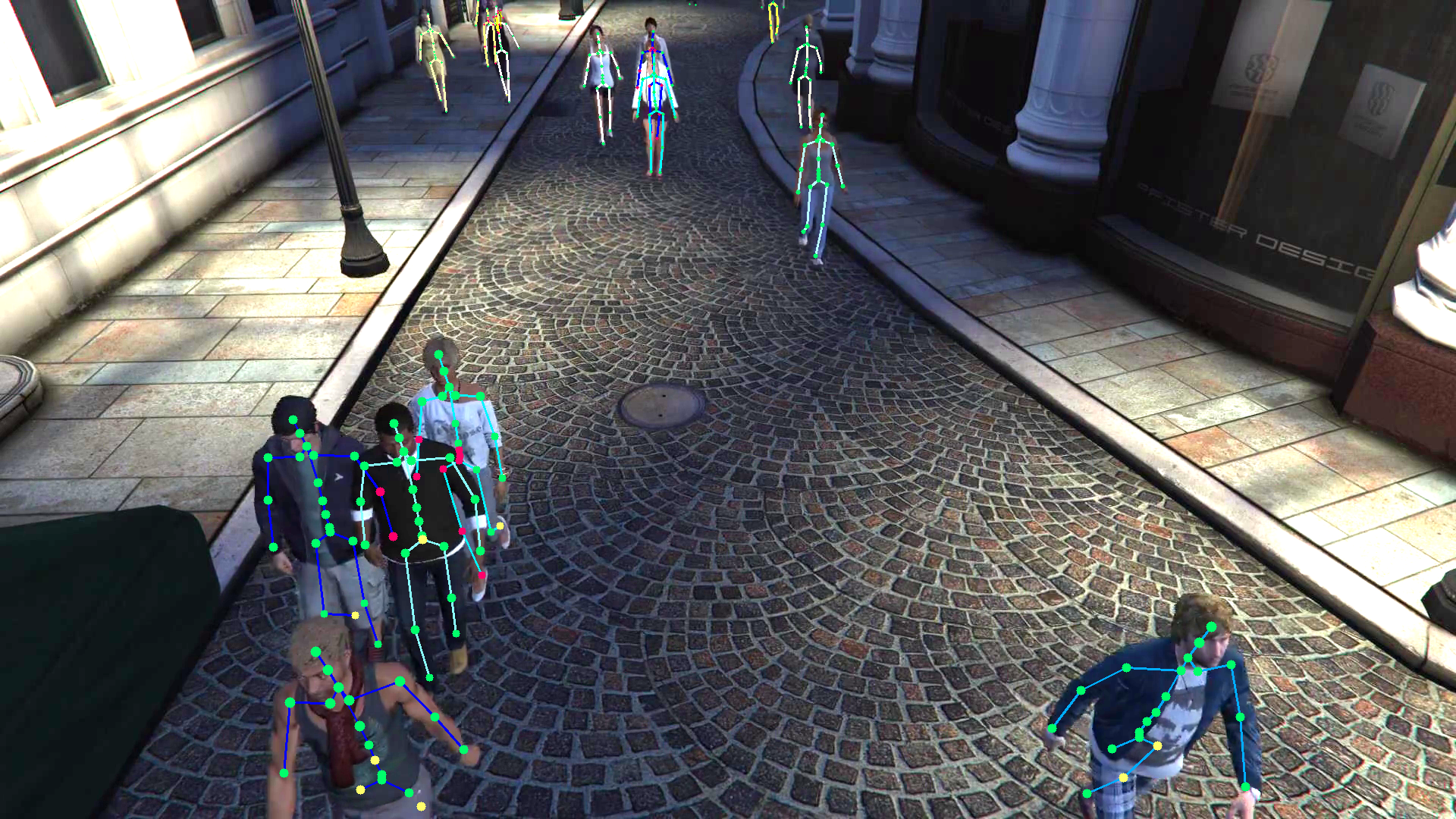}}
    \hfill
  \subfloat[\label{dataset_bbs}]{%
        \includegraphics[width=0.46\linewidth]{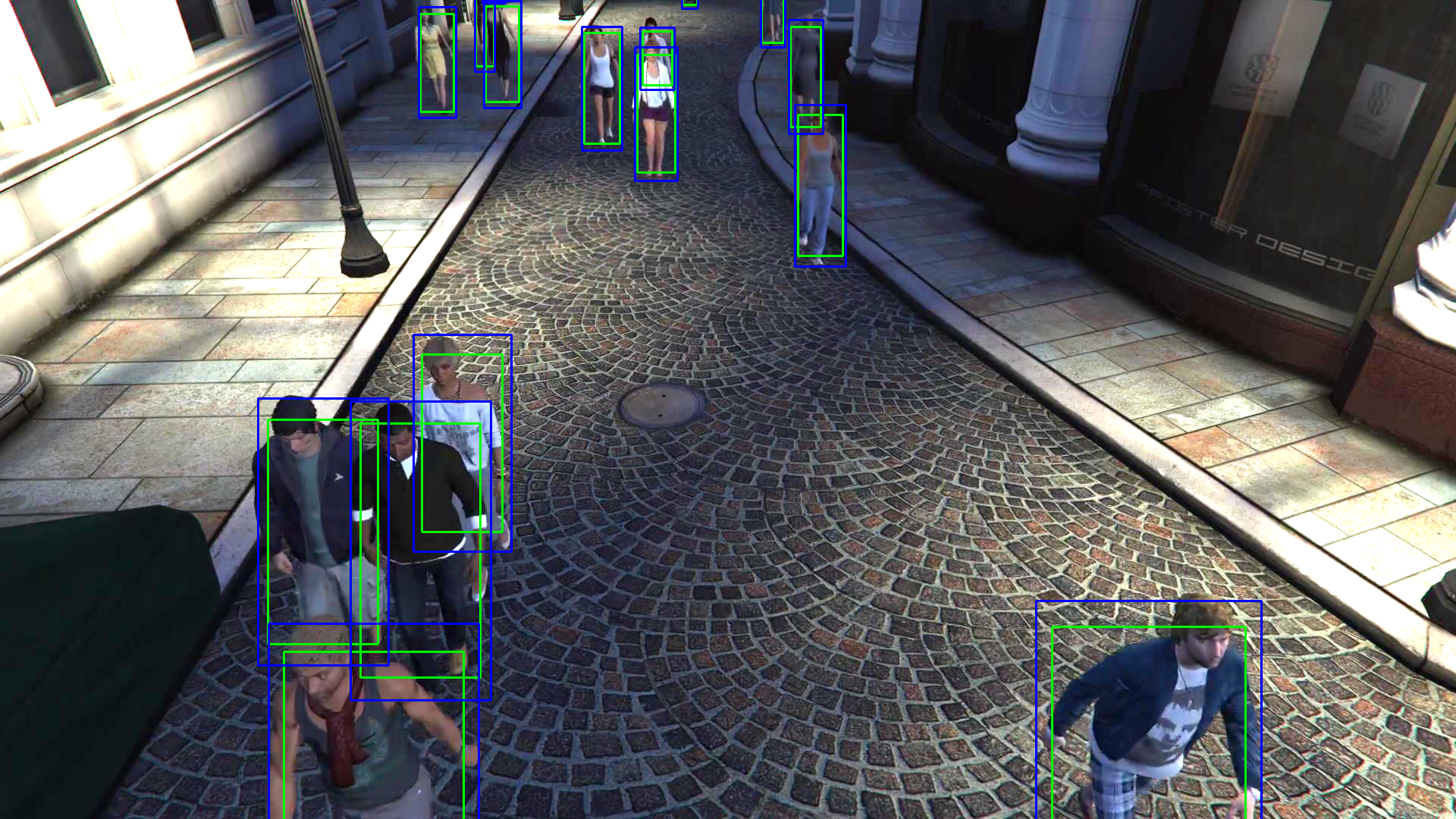}}
  \label{dataset} 
  \caption{(\textbf{a}) Pedestrians in the {JTA (Joint Track Auto)} dataset with their skeletons. (b) Examples of annotations in the \ourdataset ({Vi}rtual {Pe}destrian {D}ataset) dataset; original bounding boxes are in green, while the sanitized ones are in blue.}
\end{figure}

As a first approximation, we exploited the skeleton joints' positions in screen coordinates, directly available from the JTA annotations, for drawing the minimum bounding box enclosing all the skeleton joints (green boxes in Figure ~\ref{dataset_bbs}). However, it can be noticed that the bounding boxes produced using this simple procedure are undersized compared to the full-sized pedestrian instance, as the skeleton always lays below the skin surface. We solved this issue by constructing a bigger bounding box (blue boxes in Figure~\ref{dataset_bbs}), obtained by estimating an amount of padding through a simple heuristic.
In particular, we estimated the height of a pedestrian mesh, denoted as $h_m$, from the height $h_s$ of its skeleton, through the formula:
\begin{equation}
\label{mat:find_alpha}
    h_m = h_s + \frac{\alpha}{z}
\end{equation}
where $z$ is the distance of the pedestrian center of mass from the camera, and $\alpha$ is a parameter that depends on the camera projection matrix.

The $z$ value for each pedestrian is already included in the JTA annotations, while $\alpha$ is unknown since we can not access the camera parameters. Then, we evaluated $\alpha$ from Equation \ref{mat:find_alpha}, estimating $h_m$ for a small representative population of pedestrians. To this end, we isolated 50 random pedestrians from different scenarios, and we manually annotated them with their height in pixels units. At this point, it has been possible to recover the value of $\alpha$ from Equation \ref{mat:find_alpha} performing a simple linear regression to find the best fit.

The height padding depends basically only on the distance of the pedestrian from the camera. Instead, the width is also linked to the specific pedestrian pose. However, we found that we can ignore these pose-dependent effects while still obtaining an excellent estimate by deriving the pedestrian width $w_m$ assuming no changes in the original bounding box aspect ratio. 
For this reason, we simply derived $w_m$ from the computed $h_m$ as follows:

\begin{equation}
    w_m = h_m \frac{w_s}{h_s} = h_m r
\end{equation}
where $r$ is the aspect ratio of the bounding box enclosing the skeleton. Some examples of final estimated bounding boxes are shown in blue in Figure~\ref{dataset_bbs}.

\begin{figure}[htbp]
    \centering
    \includegraphics[width=0.62\linewidth]{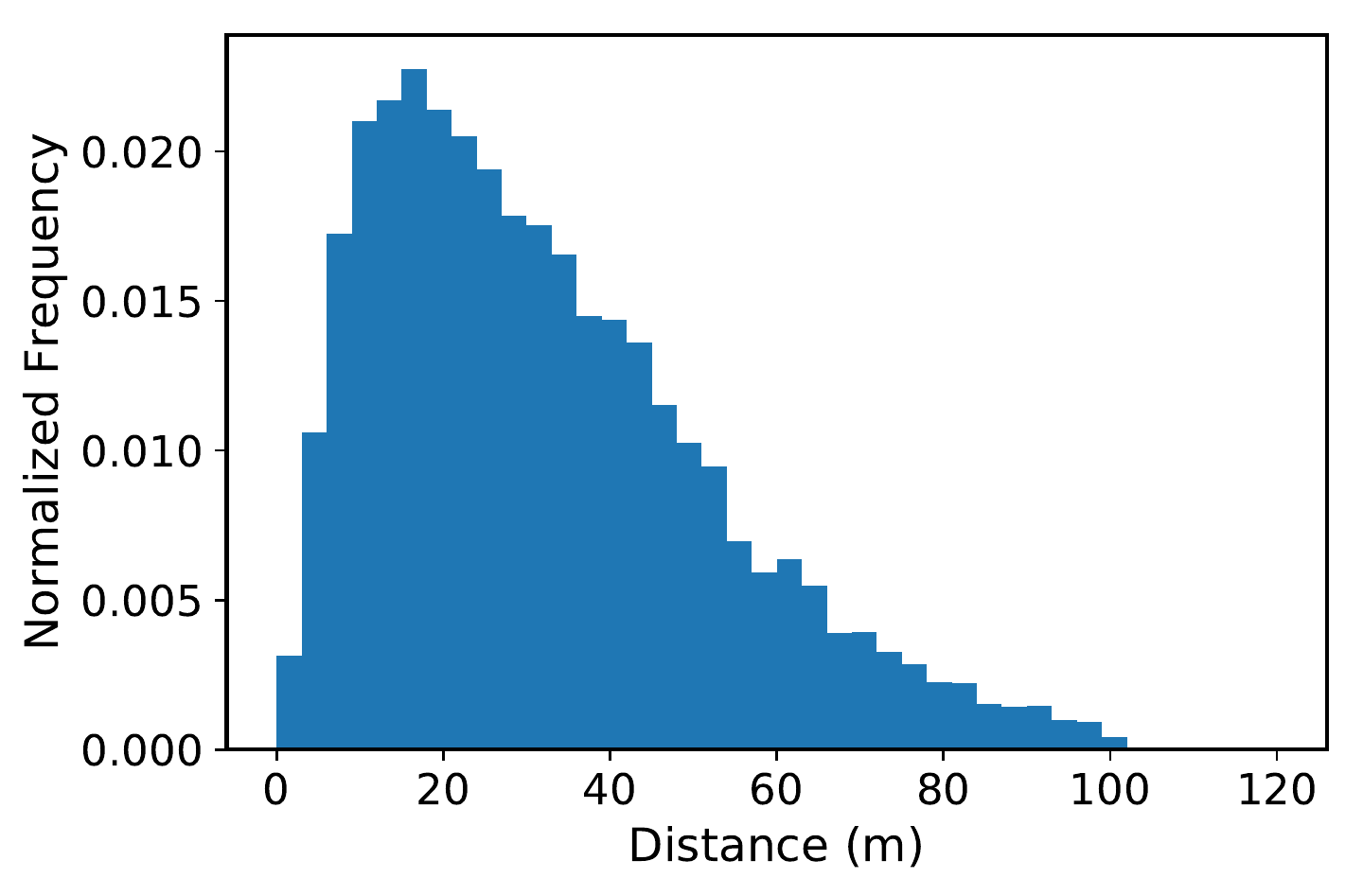}
  \caption{Histogram of distances between pedestrians and cameras.}
  \label{distance_hist} 
\end{figure}

We then assessed the quality of the produced bounding boxes. In Figure~\ref{distance_hist}, we report 
  a histogram depicting the distribution of the distances of the pedestrians from the camera. We observed that human annotators tend not to annotate pedestrians far than a certain amount from the camera in real-world datasets. We compute this distance limit by finding the minimum bounding box height, in pixels, occurring in human annotations of the MOT17Det \cite{MOT17} dataset, and seeing at what distance from the camera we reach this bounding box limit size on the JTA annotations. We concluded that human annotators do not include bounding boxes for pedestrians farther than 30--40 meters from the camera. Then, to be consistent with real-world datasets on which we will validate our approach, we cleaned the produced bounding boxes by pruning all the ones enclosing pedestrians farther than 40 meters.

\begin{figure}[htbp]
    \centering
  \subfloat[\label{1a}]{%
       \includegraphics[width=0.46\linewidth]{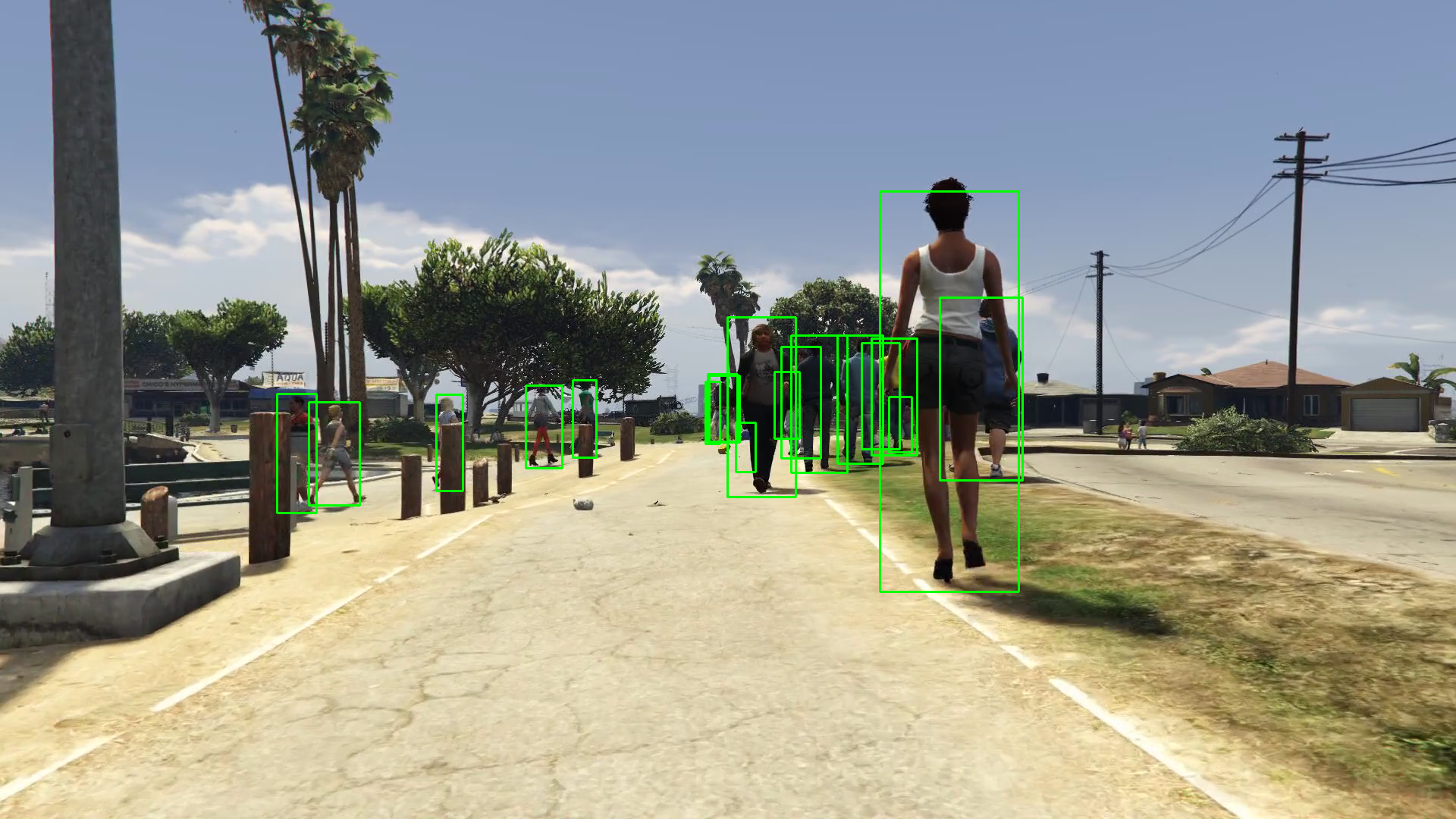}}
    \hfill
  \subfloat[\label{1b}]{%
        \includegraphics[width=0.46\linewidth]{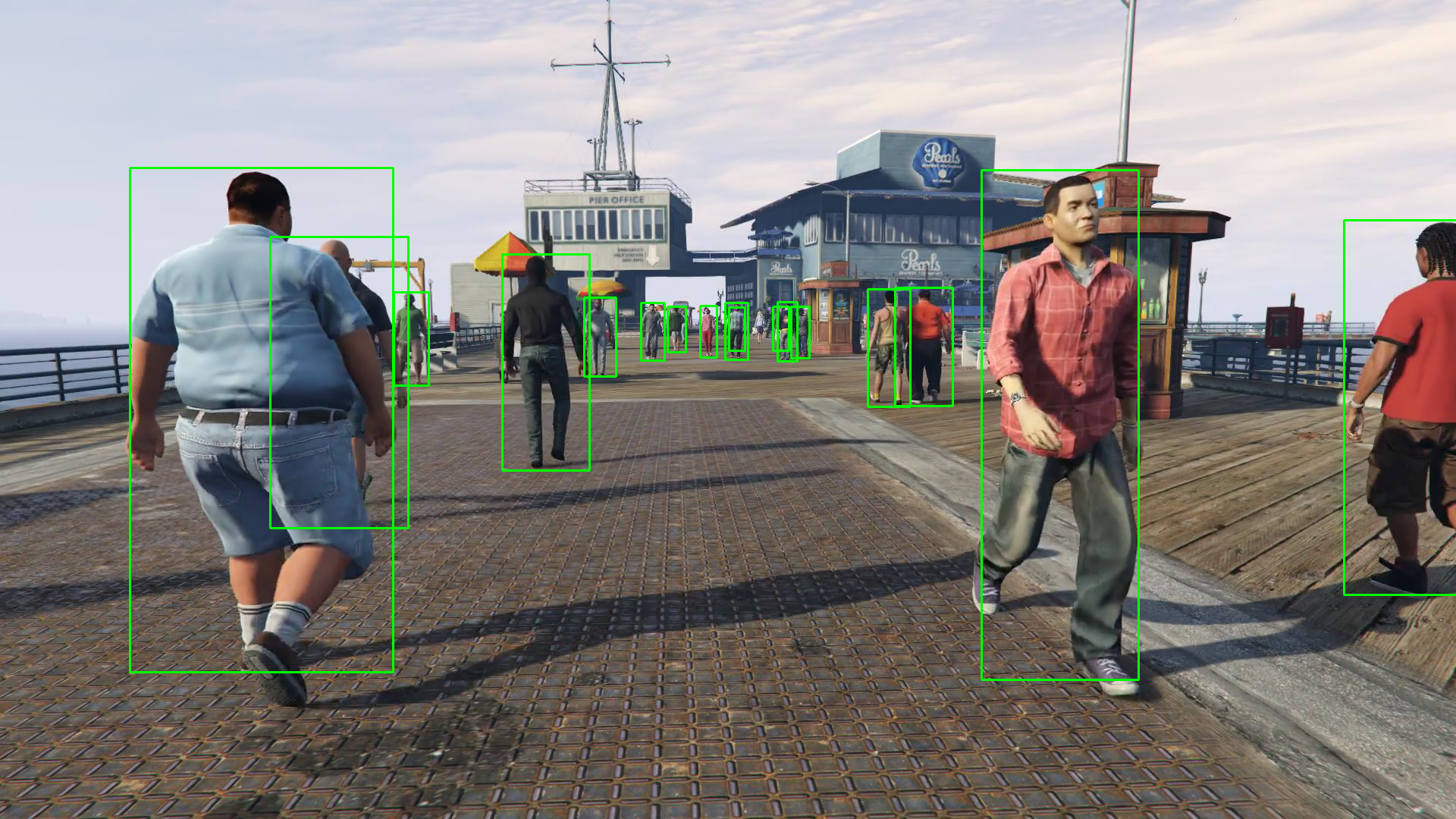}}
    \hfill
    \\
  \subfloat[\label{1d}]{%
        \includegraphics[width=0.46\linewidth]{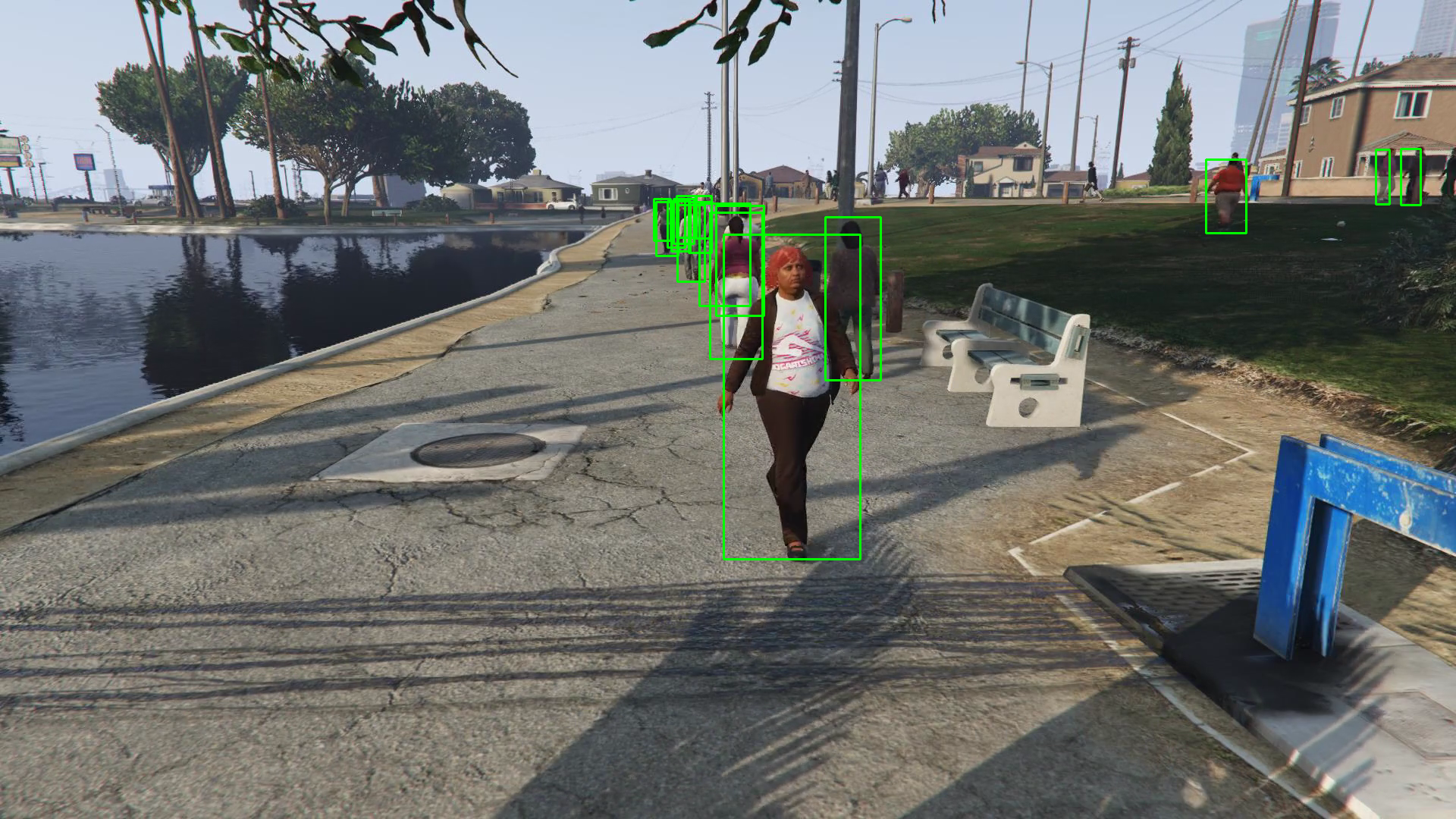}}
    \hfill
  \subfloat[\label{1e}]{%
        \includegraphics[width=0.46\linewidth]{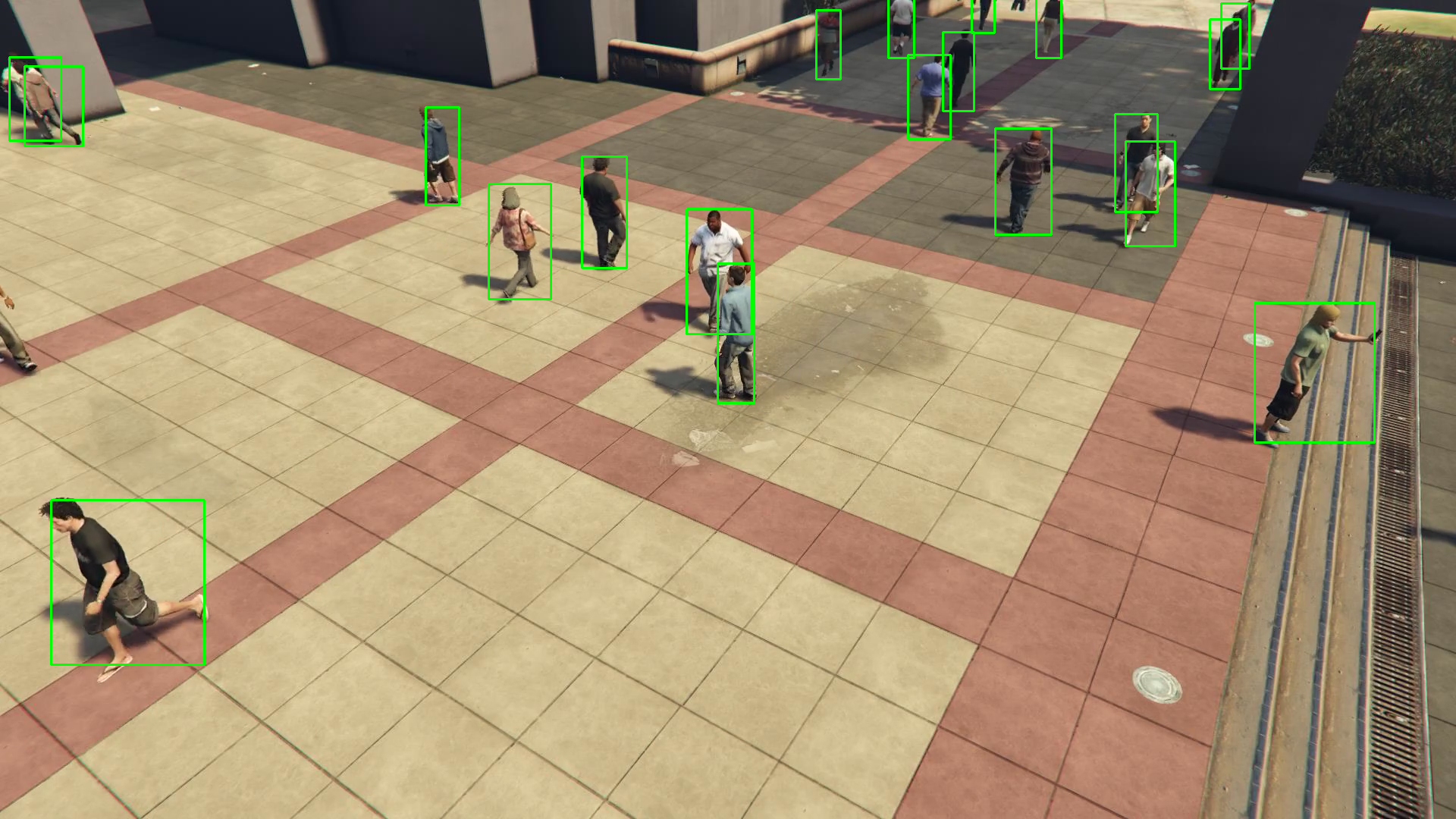}}
    \hfill
  \caption{Examples of images of the \ourdataset dataset together with the sanitized bounding boxes.}
  \label{viped_examples} 
\end{figure}

In Figure~\ref{viped_examples}, we report 
some examples of images of the \ourdataset dataset together with the sanitized bounding boxes.

\section{Domain Adaptation for Synthetic2Real Pedestrian Detection}
\label{sec:method}
In this section, we describe the object detector and the domain adaptation strategies we employed in this work. We exploited Faster R-CNN \cite{Ren2015}, a widely used state-of-the-art object detector that we briefly review in Section \ref{Faster_RCNN}.  
We train this CNN using {\ourdatasetwithoutspace}, our collection of {synthetic} images {automatically} annotated, already outlined in Section \ref{subsec:viped}. To mitigate the existing domain shift between these data and the real-world ones, we propose two domain adaptation techniques. The first one, described in Section \ref{domain_adaptation_1}, consists of training the detector with the synthetic data and then fine-tuning it exploiting the real-world images. In the second approach, described in Sec. \ref{domain_adaptation_2}, we employ instead another supervised technique, called Balanced Gradient Contribution (BGC) \cite{ros2016synthia,ros2016bgc}, where we mix the synthetic and the real-world data during the training phase. 
Figures \ref{finetuning_arch} and \ref{mixedbatch_arch} show an overview of the two solutions.

\subsection{Faster R-CNN Object Detector}
\label{Faster_RCNN}
We exploit Faster-RCNN \cite{Ren2015} as object detector architecture. In our previous work \cite{amato2019viped}, we employed YOLOv3 \cite{Redmon2018}, another state-of-the-art object detector. Here, our choice fell on Faster R-CNN since it provides better performance. Furthermore, we do not consider pedestrian detection-specific solutions since the two proposed domain adaptation techniques can also be applied to other tasks, accounting for another class of objects different from the pedestrian one.

Faster R-CNN is a two-stage CNN-based algorithm composed of different networks: The backbone, the Region Proposal Network (RPN), and the Evaluation Network (EN). In the first stage, a CNN acts as a backbone, extracting the input image features. Starting from these features space, the RPN is in charge of generating region proposals that might contain objects. Briefly, RPN slices pre-defined region boxes (called anchors) over this space and ranks them, suggesting the ones most likely containing objects. Once RPN produces the Regions Of Interests (ROIs), they might be of different sizes. Since it is hard to work on features having different sizes, RPN reduces them into the same dimension using the Region of Interest Pooling algorithm. These fixed-size proposals are finally processed by the EN, responsible for classifying and locating the objects inside them. Then, given an input image, the EN network final outputs are class scores and bounding boxes coordinates.

Faster R-CNN is then a versatile and modular network in which it is possible to change the building blocks. Regarding the backbone, our choice fell on the ResNet-50 network, a lighter version of the very popular ResNet-101 network \cite{he2016deep}. Indeed, Faster R-CNN with ResNet-50 can produce satisfactory detection results compared to the low computational resources and the time required during the training and test phases. 

\subsection{Domain Adaptation using Real-World Fine-Tuning}
\label{domain_adaptation_1}
The first proposed DA solution relies on a Transfer Learning (TL) strategy. As pointed out in \cite{wang2018deep}, DA is a particular TL case that employs labeled data in one or more relevant source domains to execute the task in a target domain. 
In particular, the crucial point in this methodology consists of fine-tuning a previously trained model with the target-domain data. 

We divide our fine-tuning methodology into two different steps.

In the first step, we consider as the baseline the Faster R-CNN detector described in the above section, having a ResNet-50 backbone pre-trained on the {COCO} dataset \cite{coco}, a large collection of images depicting complex everyday scenes of ordinary objects in their natural context, divided into 80 different categories.
Since this network is a generic object detector that can distinguish between many different classes of objects, we modify the EN building block to adapt the model to our purposes. In particular, we reduce the last fully connected layers of the detector to recognize and locate object instances belonging only to a specific category, i.e., the pedestrian category. 
Then, we train this modified Faster R-CNN-based network exploiting our synthetic images of the \ourdatasetwithoutspace, leaving all the model weights unfrozen during this phase so that the back-propagation algorithm can tune them.



\begin{figure}[htbp]
    \centering
    \includegraphics[width=0.98\linewidth]{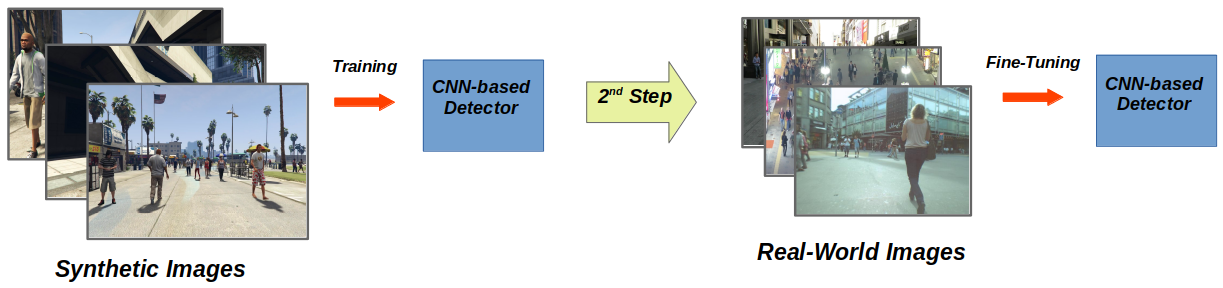}
  \caption{Overview of the first domain adaptation technique. In a first step, we train the detector using \ourdatasetwithoutspace, our synthetic collection of images. Then, in a second step, we fine-tune the network using real-world images.}
  \label{finetuning_arch} 
\end{figure}

Then, in the second step, we fine-tune this pre-trained model using real-world images as the target domain. So, in the end, the network will have processed both source and target images, memorizing in its weights information from both the domains. Figure \ref{finetuning_arch} shows 
 an overview of this approach. This method, looking at real images in this last step, is particularly useful for boosting the detector's performance on a specific real-world target scenario.

\subsection{Domain Adaptation using Balanced Gradient Contribution}
\label{domain_adaptation_2}
The second DA approach is an end-to-end training, so it benefits from not relying on a two-step process like the previous one.

As in the previous solution, we start with the modified Faster R-CNN detector having the ResNet-50 backbone pre-trained on the {COCO} dataset. This time, we train the network using mixed batches, i.e., we employ batches containing synthetic and real-world images simultaneously, given a fixed mixing ratio. As explained in \cite{ros2016bgc}, the real-world data acts as a regularization term over the synthetic data training loss. In particular, we exploit batches composed of $2/3$ of synthetic images and of $1/3$ of real-world data. Thus, statistics from both domains are considered throughout the entire procedure, creating a more accurate model for both.

\begin{figure}[htbp]
    \centering
    \includegraphics[width=0.95\linewidth]{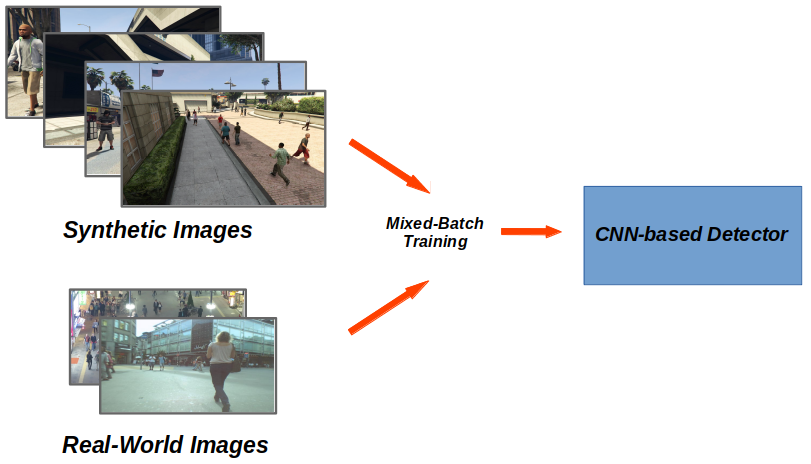}
  \caption{Overview of the second domain adaptation technique. We mitigate the {Synthetic2Real} domain shift in a single-step training procedure, employing mixed batches containing both synthetic and real images at the same time.}
  \label{mixedbatch_arch} 
\end{figure}

Again, during this phase, we leave all the network weights unfrozen so that the back-propagation algorithm can modify the network parameters accordingly. Consequently, we mitigate the {Synthetic2Real} Domain Shift straight in a single-step training. Figure \ref{mixedbatch_arch} shows an overview of this approach.

In the experiments, we employed this technique for boosting the detector's performance on a precise target scenario, using batches composed of the synthetic data and the real-world images specific for the particular considered scenes. Besides, we exploited this solution also for achieving wide generalization capabilities, considering batches containing synthetic images and generic real-world images containing pedestrians.

\section{Experimental Evaluation}
In this section, we briefly report some details about the real-world datasets exploited for the experiments. Then, we show and discuss the results concerning the generalization capabilities of our detector trained using \ourdatasetwithoutspace. Finally, we illustrate the performance of the two domain adaptation techniques over specific real-world scenarios.

\subsection{Real-World Datasets}
\label{sec:mot_datasets}

{MOT17Det} \cite{MOT17} and {MOT19Det} \cite{MOT19} are newly introduced datasets with manual annotations for pedestrian detection that are particularly suitable for surveillance applications. They comprise a collection of challenging images (5,316 and 8,931, respectively) taken from multiple sequences with various crowded scenarios having different viewpoints, weather conditions, and camera motions. 
The authors provided training and test subsets, but they released only the ground-truth annotations belonging to the former. The performance metrics concerning the test subsets are instead available submitting results to their {MOT Challenge} website ({\url{https://motchallenge.net}}).
The main peculiarity of {MOT19Det} compared to {MOT17Det} is the massive crowding of the collected scenarios.

{CityPersons} dataset \cite{citypersons} consists of a large and diverse set of stereo video sequences recorded in streets from different cities in Germany and neighboring countries. In particular, the authors provide 5,000 images from 27 cities labeled with bounding boxes and divided across train/validation/test subsets. This dataset is more focused on self-driving applications, and images are collected from a moving car.

{COCOPersons} dataset is a split of the popular {COCO} dataset \cite{coco} comprising images collected in general contexts belonging to 80 categories. We filter these images considering only the ones belonging to the {persons} category. Hence, we obtain a new dataset of about 66,000 images containing at least one pedestrian instance.

\subsection{Experiments}
We evaluate the detection performances using the standard mean average precision (mAP) metric. In particular, we consider the detection proposals having a score confidence greater than 0.05. Then, we employ the COCO mAP \cite{coco} and the MOT AP metrics \cite{MOT17}, fixing the IoU threshold to 0.5 and varying only the detection confidence threshold.

\subsubsection{Testing Generalization Capabilities}
To test the generalization capabilities, we train the detector on a source domain, and then we validate it on a different target domain. In particular, we train the model using a dataset, and then we test it on another one. In this way, we guarantee that the two distributions are different and not related.

In particular, we train the modified Faster R-CNN-based detector described in Section \ref{Faster_RCNN} using \ourdatasetwithoutspace. This procedure corresponds to the first step of the previously described domain adaptation solution (see \ref{domain_adaptation_1}). We evaluate this model testing it on the real-world datasets {MOT17Det}, {MOT19Det} and {CityPersons}, defining three validation subsets containing images not present in the training subset.

To form a solid baseline for this experiment, we train the same detector using every one of the three real-world datasets, and then we test them over the remaining two. We also report a further baseline considering the detector trained only on the real-world general-purpose {COCO} dataset, considering only the detections belonging to the {person} category.

We also experiment with the mixed-batch domain-adaptation approach explained in Section \ref{domain_adaptation_2}, using the same evaluation protocol as before. We exploit batches composed of $2/3$ of \ourdataset and by the remaining $1/3$ of {COCOPersons}. We choose the latter as the real-world dataset since it depicts humans in highly heterogeneous scenarios, and it is not biased towards a specific application (e.g., autonomous driving). Again, we evaluate this model testing on all the three remaining real-world datasets.

We report the results in Table~\ref{tab1}. 
Note that we omit results concerning a specific dataset if employed during the training phase, for a fair evaluation of the overall generalization capabilities.

\begin{table}[htbp]
\caption{Evaluation of the generalization capabilities. The first section of the table reports results obtained training the detector with real-world data, while the latter is related to the model trained over synthetic images. \ourdataset + Real refer to the mixed batch experiments with $2/3$ ViPeD and $1/3$ of COCOPersons. Results are evaluated using the COCO mAP. We report in bold the best results.}
\begin{center}
\begin{tabular}{lccc}
\toprule
& \multicolumn{3}{c}{\textbf{Test Dataset}} \\
\cmidrule(lr){2-4}
\textbf{Training Dataset} & MOT17Det & MOT19Det & CityPersons \\
\midrule
COCO & 0.636 & 0.466 & 0.546 \\
\midrule
MOT17Det & - & 0.605 & \textbf{0.571} \\
\midrule
MOT19Det & 0.618 & - & 0.419 \\
\midrule
CityPersons & 0.710 & 0.488 & - \\
\midrule
\ourdataset & 0.721 & \textbf{0.629} & 0.516 \\
\midrule
\ourdataset + Real & \textbf{0.733} & 0.582 & 0.546 \\
\bottomrule
\end{tabular}
\label{tab1}
\end{center}
\end{table}

In most cases
 , as we can see, our network performs better than those trained using only the manually annotated real-world datasets, taking advantage of the high variability and size of the \ourdataset dataset. In particular, concerning the {MOT17Det} dataset, all our solutions trained with synthetic data outperform those trained with real ones. We obtain the best results using the mixed-batch approach. Considering the {MOT19Det} dataset, we achieve the best result in training the detector with \ourdatasetwithoutspace. {CityPersons} is the only dataset on which the algorithm maintains higher performances when trained with real-world data. In particular, the highest mAP on {CityPersons} is obtained when the detector is trained with the {MOT17Det} dataset. However, the mixed-batch approach achieves, in this case, results comparable with the baselines.

\subsubsection{Testing Domain Adaptation Techniques over Specific Real-world Scenarios}
To test how the two proposed domain adaptation techniques behave when considering specific target real-world scenarios, we consider the {MOT17Det} and {MOT19Det} real-world datasets.

Regarding the fine-tuning DA approach, we consider as training sets those proposed by the authors of \cite{MOT17, MOT19}, and we obtain the evaluation of our results over the test sets submitting them to the {Mot Challenge} website. For the mixed-batch DA solution, during the training phase, we inject in the same batch $2/3$ of synthetic images from the \ourdataset dataset and $1/3$ of real-world images from the training subsets of the {MOTDet17} or the {MOT19Det} dataset. Again, we validate our results by submitting them to the {Mot Challenge} website.

Table~\ref{tab2} and Table~\ref{tab3} report the results for the two considered scenarios. 
We report our results together with the state-of-the-art approaches publicly released in the MOT Challenges (at the time of writing).

\begin{table}[htbp]
\caption{Evaluation of the two {Domain Adaptation} (DA) techniques on the {MOT17Det} dataset. FT-DA (Fine Tuning DA) is the first proposed solution, while MB-DA (Mixed Batch DA) is the second one. Results are evaluated using the MOT mean average precision (mAP).}
\begin{center}
\begin{tabular}{ccc}
\toprule
\textbf{Method} & \textbf{MOT AP} \\
\midrule
YTLAB \cite{Cai2016} & 0.89 \\
\midrule
KDNT \cite{poi} & 0.89 \\
\midrule
ViPeD FT-DA (our) & \textbf{0.89} \\
\midrule
ViPeD MB-DA (our) & 0.87\\
\midrule
ZIZOM \cite{Lin2018} & 0.81 \\
\midrule
SDP \cite{Yang2016} & 0.81 \\
\midrule
FRCNN \cite{Ren2015} & 0.72 \\
\bottomrule
\end{tabular}
\label{tab2}
\end{center}
\end{table}

\begin{table}[htbp]
\caption{Evaluation of the two DA techniques on the {MOT19Det} dataset. FT-DA (Fine Tuning DA) is the first proposed solution, while MB-DA (Mixed Batch DA) is the second one. Results are evaluated using the MOT mAP.}
\begin{center}
\begin{tabular}{ccc}
\toprule
\textbf{Method} & \textbf{MOT AP} \\
\midrule
SRK\_ODESA & \textbf{0.81} \\
\midrule
CVPR19\_det & 0.80 \\
\midrule
Aaron & 0.79 \\
\midrule
PSdetect19 & 0.74 \\
\midrule
ViPeD FT-DA \cite{amato2019viped} (our) & 0.80 \\ 
\midrule
ViPeD MB-DA \cite{amato2019viped} (our) & 0.80 \\
\bottomrule
\end{tabular}
\label{tab3}
\end{center}
\end{table}


As we can see, the two DA approaches can mitigate the {Synthetic2Real} Domain Shift. In both datasets, we obtain an improvement in performance compared to the results in Table \ref{tab1}. It is also worth noting that we achieve competitive results in both scenarios compared to the state-of-the-art, reaching the first and the second places in the leader boards of the {MOT17Det} and {MOT19Det} challenges, respectively.

\section{Conclusions}

In this work, we addressed the pedestrian detection task by proposing a CNN-based solution trained using synthetically generated data. The choice of training a CNN using synthetic data is motivated by the fact that the network, to generalize well, requires a considerable amount of manually annotated images representing different scenarios. This procedure usually requires a significant human effort, and it is error-prone. 

To this end, we introduced a synthetic dataset named {\ourdataset}, containing a massive collection of images rendered from the highly photo-realistic video game {GTA V} developed by {Rockstar North} and a full set of precise bounding boxes annotations around all the visible pedestrians. To the best of our knowledge, it is the first synthetic dataset suitable for the pedestrian detection task.

Furthermore, we proposed two different Domain Adaptation techniques to mitigate the {Synthetic2Real} Domain Shift, which are suitable for the pedestrian detection task and possibly applicable to more general object detection tasks.

The experiments showed that, in most cases, the detector trained with the synthetic data can generalize better on unseen scenarios than the same algorithm trained using only the manually annotated real-world datasets. Moreover, the two proposed DA approaches can mitigate the underlying differences between the two worlds, obtaining a performance improvement on specific real-world scenarios. 

In our opinion, the result of this work opens new perspectives to address the scalability of pedestrian and object detection methods for large physical systems with limited supervisory resources. Using our freely available model trained using {\ourdataset}, future researchers will have at their disposal a detector able to localize instances of people over images belonging to a multitude of different scenarios and, therefore, a system robust to newly added sources of data. On the other hand, they will also have the possibility of further specializing the detector to work over new added real-world scenarios using our two domain adaptation techniques, obtaining an additional performance boost.

\bibliographystyle{unsrt}  
\bibliography{references}

\end{document}